\documentclass[runningheads]{llncs}
\usepackage{graphicx}
\usepackage{comment}
\usepackage{amsmath,amssymb} 
\usepackage{color}

\usepackage{bold-extra}

\newcommand{\ganh}{{\sc GANHopper}}
\DeclareMathOperator{\Lagr}{\mathcal{L}}
\newcommand{\rz}[1]{{\color{black}#1}}
\newcommand{\rzz}[1]{{\color{black}#1}}
\newcommand{\wm}[1]{{\color{black}#1}}

\usepackage{algorithm2e}
\begin{document}
\title{\textsc{\textbf{GANHopper}}: Multi-Hop GAN for Unsupervised Image-to-Image Translation}
\titlerunning{\textsc{GANHopper}}
%
\author{Wallace Lira\inst{1} \and
Johannes Merz\inst{1} \and 
Daniel Ritchie\inst{2} \and \\
Daniel Cohen-Or\inst{3} \and
Hao Zhang\inst{1}}
\authorrunning{W. Lira et al.}
%
\institute{Simon Fraser University, Canada\\
\email{\{wpintoli, johannes\_merz, haoz\}@sfu.ca}\\
\and
Brown University, USA\\
\email{daniel\_ritchie@brown.edu}
\and
Tel Aviv University, Israel\\
\email{dcor@tau.ac.il}
}
\maketitle              

\begin{figure}[] \centering
    \includegraphics[width=0.99\linewidth]{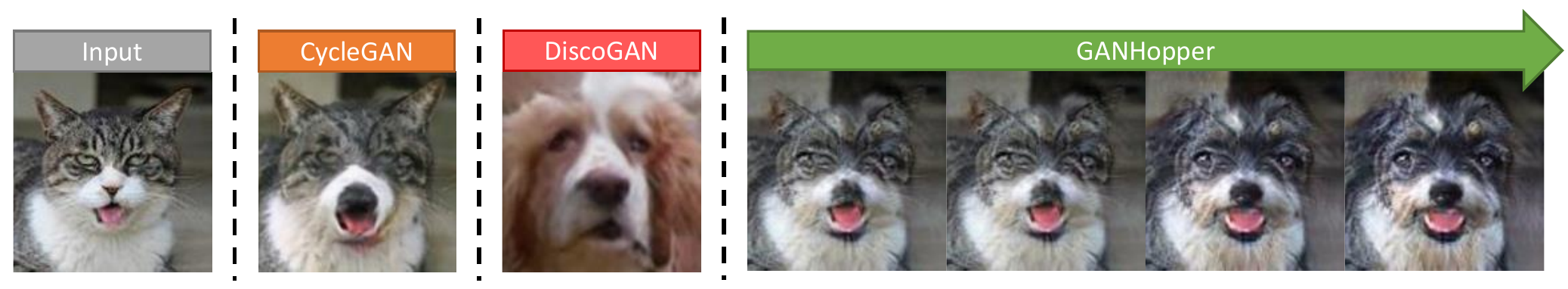}
    \caption{{
    \em What dog would look most similar to a given cat?\/} Our {\em multi-hop\/} image translation network, \ganh, produces such transforms and also {\em in-between\/} transitions through ``hops''.
    Direct translation methods can ``undershoot the target'' by failing to produce the necessary geometry variations (middle left) 
    or ``overshoot'' by significantly altering non-domain-specific features such as general color schemes (middle right). 
    }
    \label{fig:teaser}
\end{figure}

\begin{abstract}
We introduce \ganh, an unsupervised image-to-image translation network that transforms images {\em gradually\/} between two domains, through multiple {\em hops\/}. Instead of executing translation directly, we steer the translation by requiring the network to produce {\em in-between\/} images that resemble {\em weighted hybrids\/} between images from the input domains.
Our network is trained on {\em unpaired\/} images from the two domains only, without any in-between images. All hops are produced using a {\em single generator\/} along each direction. In addition to the standard cycle-consistency and adversarial losses, we introduce a new {\em hybrid discriminator\/}, which is trained to classify the intermediate images produced by the generator as weighted hybrids, with weights based on a predetermined hop count. 
We also add a smoothness term to constrain the magnitude of each hop, further regularizing the translation. Compared to previous methods, \ganh~excels at image translations involving domain-specific image features and geometric variations while also preserving non-domain-specific features such as general color schemes.
\end{abstract}

\section{Introduction}
\label{sec:intrp}

Unsupervised image-to-image translation has been one of the most intensively studied problems in computer vision, since the introduction of domain transfer network (DTN)~\cite{DTN}, CycleGAN~\cite{zhu2017CycleGAN}, DualGAN~\cite{DualGAN},
and UNIT~\cite{UNIT} in 2017. While these networks and many follow-ups were designed to perform general-purpose translations, it is challenging for the translator to learn transformations beyond local and stylistic adjustments, such as geometry and shape variations. For example, typical dog-cat translations learned by CycleGAN do not transform the animals in terms of geometric facial features; only pixel-scale color or texture alterations take place, e.g., see Figure~\ref{fig:teaser} (middle right).

When the source and target domains exhibit sufficiently large discrepancies, any proper translation function is expected to be complex and difficult to learn.
Without any paired images to supervise the learning process, the search space for the translation functions can be immense. With large image changes, there are even more degrees of freedom to account for.
In such cases, a more {\em constrained\/} and {\em steerable\/} search would be desirable.

In this paper, we introduce an unsupervised image-to-image translator that is constrained to transform images {\em gradually\/} between two domains, e.g., cats and dogs.
Instead of performing the transformation directly, our translator executes the task in steps, called {\em hops\/}. Our {\em multi-hop\/} network is built on 
CycleGAN~\cite{zhu2017CycleGAN}. However, we steer the translation paths by forcing the network to produce {\em in-between\/} images which resemble {\em weighted hybrids\/} between images from the two input domains. For example,
a four-hop network for dog-to-cat translation produces three in-between images: the first is 25\% cat-like and 75\% dog-like, the second is 50/50 in terms of cat and dog likeness, and the third is 75\% cat-like and 25\% dog-like.
The fourth and final hop is a 100\% translated cat.

Our network, \ganh, is unsupervised and trained on unpaired images from two input domains, without any in-between hybrid images in its training set. Equally important, all hops are produced using a {\em single generator\/} along each direction, hence the network capacity does not exceed that of CycleGAN. To make training possible, we introduce a new {\em hybrid discriminator\/}, which is trained exclusively on real images (e.g., dogs or cats) to evaluate the in-between images by classifying them as weighted hybrids, depending on the prescribed hop count. In addition to the original cycle-consistency and adversarial losses from CycleGAN, we introduce two new losses: a {\em hybrid loss\/} to assess the degree to which an image belongs to one of the input domains, and a {\em smoothness\/} loss which further regulates the image transitions to ensure that a generated image in the hop sequence does not deviate much from the preceding image. 

\ganh~does not merely transform an input cat into {\em a\/} dog --- many dogs can fool the discriminator. Rather, it aims to generate {\em the\/} dog which looks most similar to the given cat; see Figure~\ref{fig:teaser} (middle left).
Compared to previous unsupervised image-to-image translation networks, our network excels at image translations
involving domain-specific image features and geometric variations (i.e., ``what makes a dog a dog?")
while preserving non-domain-specific image features such as the fur color of the input cat in Figure~\ref{fig:teaser}.

The ability to produce large changes, in particular, geometry transformations, via unsupervised domain translation has been a hotly-pursued problem. There appears to be a common belief that the original CycleGAN/DualGAN architecture cannot learn geometry variations. To do so, the feature representation and/or training approach must be changed. As a result, many approaches resort to latent space translations, e.g., with style-content~\cite{MUNIT} or scale~\cite{LOGAN} separation and feature disentanglement~\cite{TransGaGa}.
Our work challenges this assumption, as \ganh~follows fundamentally the same architecture as CycleGAN, working directly in image space;
it merely enforces a gradual, multi-hop translation to steer and regulate the image transitions.
\rz{In addition, multi-hop GANs represent a generic ``meta idea'' which is quite extensible, e.g., in terms of varying the number and architecture of the in-between translators. As demonstrated by Figure~\ref{fig:teaser} and more results later, even the simplest option of using {\em one\/} network can already make a significant difference for various domain translation tasks.}


%

\section{Related Work}

The foundation of modern image-to-image translation is the UNet architecture, first developed for semantic image segmentation~\cite{UNet}.
This architecture was later extended with conditional adversarial training to a variety of image-to-image translation tasks~\cite{IsolaZZE16}.
Further improvements led to the generation of higher-resolution outputs~\cite{Pix2PixHD} and multiple possible outputs for the same image in ``one-to-many'' translation tasks, e.g. grayscale image colorization~\cite{BicycleGAN}.

The above methods require paired input and output images $\{(x_i, y_i)\}$ as training data.
A more recent class of image-to-image translation networks is capable of learning from only \emph{unpaired} data in the form of two sets $\{x_i\}$ and $\{y_i\}$ of input and output images, respectively~\cite{zhu2017CycleGAN,DualGAN,DiscoGAN}.
These methods jointly train a network $G$ to map from $x$ to $y$ and a network $F$ to map from $y$ to $x$, enforcing at training time that $F(G(x)) = x$ and $G(F(y)) = y$.
Such \emph{cycle consistency} is thought to regularize the learned mappings to be semantically meaningful, rather than arbitrary translations.

While the above approaches succeed at domain translations involving low-level appearance shift (e.g. summer to winter, day to night), they often fail when the translation requires a significant shape deformation (e.g. cat to dog).
Cycle-consistent translators have been shown to perform larger shape changes when trained with a discriminator and perceptual loss function that consider more global image context~\cite{Gokaslan2018}.
An alternative approach is to interpose a shared latent code $z$ from which images in both domains are generated (i.e. $x = F(z)$ and $y = G(z)$)~\cite{UNIT}.
This method can also be extended to enable translation into multiple output images~\cite{MUNIT}.
Another tactic is to explicitly and separately model geometry vs. appearance in the translation process.
A domain-specific method for translating human faces to caricature sketches accomplishes this by detecting facial landmarks, deforming them, then using them to warp the input face~\cite{CariGAN}.
More recent work has proposed a related technique that is not specific to faces~\cite{TransGaGa}.
Finally, it is also possible to perform domain translation via the feature hierarchy of a pre-trained image classification network~\cite{CascadedDeepFeatureTranslation}.
This method can also produce large shape changes.

In contrast to the above, we show that direct image-to-image translation can produce large shape changes, while also preserving appearance details, if translation is performed in a sequence of smooth hops.
This process can be viewed as producing an interpolation sequence between two domains.
Many GANs can produce interpolations between images via linear interpolation in their latent space.
These interpolations can even be along interpretable directions which are either specified in the dataset~\cite{FaderNetworks} or automatically inferred~\cite{InfoGAN}.
However, GAN latent space interpolation does not perform cross-domain interpolation. 
\rzz{Aberman et al.~\cite{NeuralBestBuddies} perform cross-domain interpolation by identifying corresponding points on images from two domains and using these points as input to drive image morphing~\cite{AutomatingImageMorphing}.
However, this approach requires images in both the source and target domain to interpolate between, whereas our method takes just a source image and produces an interpolation to the best-matching target image.
Lastly, InstaGAN~\cite{instagan} addresses large shape changes, e.g., pants to skirts, by using a multi-instance transfiguration network, relying on segmentation masks to translate one instance at a time. Their implementation includes a sequential minibatch inference/training option for a different purpose: to alleviate GPU overload when translating a large number of instances.}

\section{Method}
\label{sec:method}

\begin{figure*}[!t]
    \centering
	\includegraphics[width=0.9\textwidth]{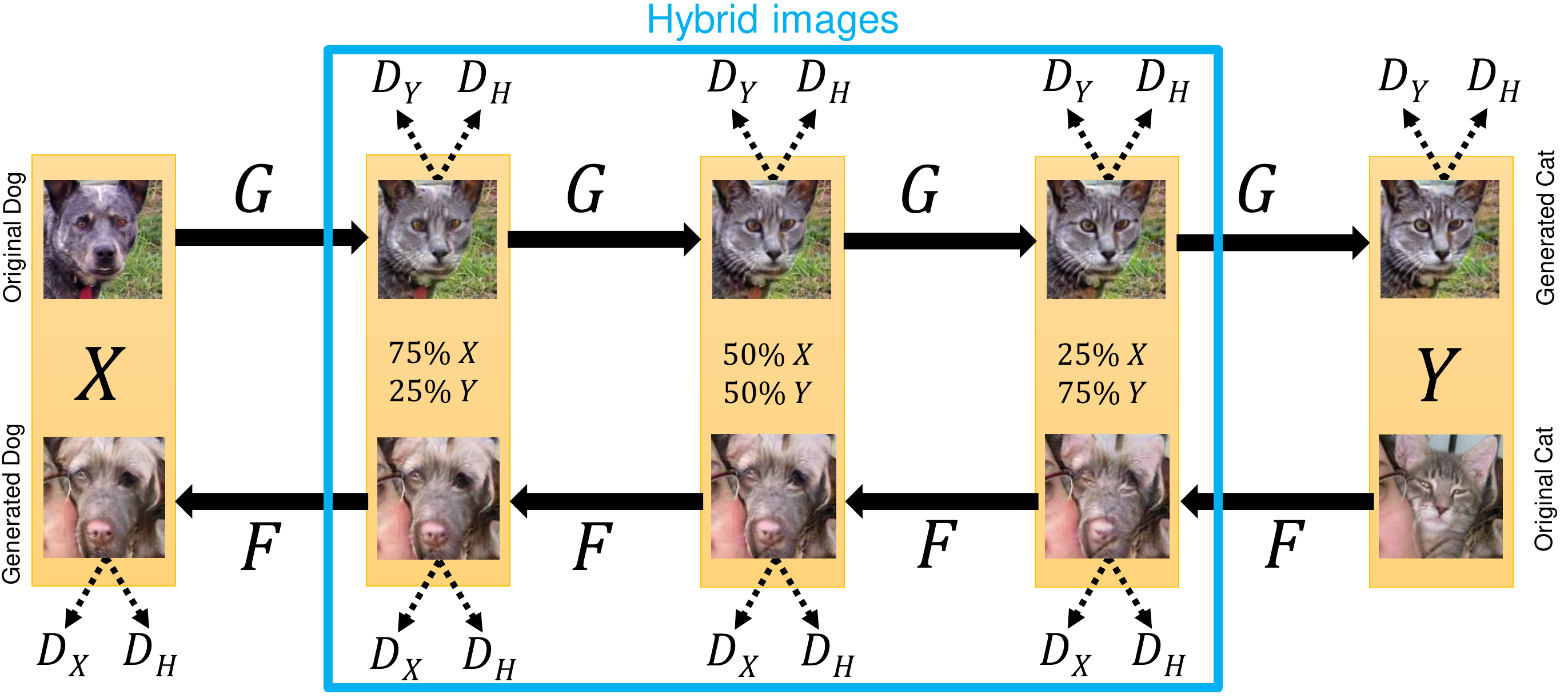}
	\caption{Let $X$ and $Y$ represent two domains that we wish to translate (dogs and cats, respectively, in this figure). Our approach warps images from $X$ to $Y$ using the generator $G$ and from $Y$ to $X$ using the generator $F$ by applying each generator $h$ times. The generator is trained by combining: \textbf{(a)} the adversarial loss, obtained by feeding the generated images, including the \textit{hybrid images}, to either $D_X$ (from $X$ to $Y$) or $D_Y$ (from $Y$ to $X$); \textbf{(b)} the reconstruction loss, which is the result of comparing a generated image, including \textit{hybrid images}, or input $i$ with either $G(F(i))$ if $i$ is being translated from $X$ to $Y$ or $F(G(i))$ if $i$ is being translated from $Y$ to $X$; \textbf{(c)} a domain \textit{hybrid loss}, a membership score to either class determined by evaluating every generated image with the hybrid discriminator $D_H$, which is trained exclusively on real images to classify the input as being either a member of $X$ or $Y$.}
	\label{fig:cycle_interpolation}
\end{figure*}

Let $X$ and $Y$ denote our source and target image domains, respectively.
Our goal is to learn a transformation that, given an image $x \in X$, outputs another image $y' \in Y$ such that $y'$ is perceived to be the counterpart of the image $x$ in the dataset $Y$. The same must be achieved with the analog transformation from $y \in Y$ to $x' \in X$. This task is identical to that performed by CycleGAN~\cite{zhu2017CycleGAN}.
However, we do not translate the input image in one pass through the network.
Rather, we facilitate the translation process via a sequence of intermediate images.
We introduce the concept of a \textbf{hop}, which we define as the process of warping one image toward the target domain by a limited amount using a generator network.
Repeated hops produce \textit{hybrid images} as byproducts of the translation process.

Since we do not translate images in a single pass through a network,
our training process must be modified from the traditional cycle-consistent learning framework.
In particular, the generation of \textit{hybrid images} during the translation is a challenge,
because the training data does not include such images. Therefore, the \textit{hybridness} of these generated images
must be estimated on the fly during training.
To this end, we introduce a new discriminator, which we call the \textit{hybrid discriminator}, whose objective is to evaluate how similar an image is to both input domains, generating a membership score.
We also add a new \textit{smoothness term} to the loss, whose purpose is to encourage a gradual warping of the images through the hops so that the generator does not overshoot the translation. The following subsections present the multi-hop framework.


\subsection{Multi-hop framework}

Our model consists of the original two generators from CycleGAN, denoted by $G$ and $F$, and three discriminators, two of which are CycleGAN's original adversarial discriminators $D_Y$ and $D_X$.
The third discriminator is the new \textit{hybrid discriminator} $D_H$.
Figure~\ref{fig:cycle_interpolation} depicts how these different generators and discriminators work together during training time to translate images via multiple hops.

\paragraph{Hop nomenclature.} A \textbf{hop} is defined as using either $G$ or $F$ to warp an image towards the domain $Y$ or $X$, respectively. A full translation is achieved by performing $h$ hops using the same generator, where $h$ is a user defined value. For instance, if $h=3$, $G(G(G(x)))=y'$, where $x \in X$ and $y' \in Y$. Similarly, $F(F(F(y)))=x'$, where $y \in Y$ and $x' \in X$. Given an image $i$, the translation hops are defined via the following recurrence relations:

\begin{equation}
\begin{aligned}
    G_h(i) &= G(G_{h-1}(i)) \qquad G_0(i) = i
    \\
    F_h(i) &= F(F_{h-1}(i)) \qquad F_0(i) = i
\end{aligned}
\end{equation}


\paragraph{Generator architecture.}
We adopt the architecture and layer nomenclature originally proposed by Johnson et al.~\cite{JohnsonAL16} and used in CycleGAN. Let c7s1-$k$ denote a 7$\times$7 Convolution-InstanceNorm-ReLU layer with $k$ filters and stride $1$. d$k$ denotes a 3$\times$3 Convolution-InstanceNorm-ReLU layer with $k$ filters and
stride 2. Reflection padding was used to reduce artifacts. R$k$ denotes a residual block with two 3$\times$3 convolutional layers, each with $k$ filters.
u$k$ denotes a 3$\times$3 TransposeConvolution-InstanceNorm-ReLU layer with $k$ filters and stride $1/2$. The network takes 128$\times$128 images as input and consists of the following layers:
c7s1-64, d128, d256, R256 ($\times$12), u128, u64, c7s1-3.

\paragraph{Discriminator architecture.}
For the discriminator networks $D_Y$, $D_X$ and $D_H$, we use the same 70$\times$70 PatchGAN~\cite{IsolaZZE16} used in CycleGAN. Let C$k$ denote a 4$\times$4 Convolution-InstanceNorm-LeakyReLU layer with $k$ filters and stride $2$.
Given the 128$\times$128 input images, we produce a 16$\times$16 feature matrix. Each of its elements is associated with one of the 70$\times$70 patches from the input image. The discriminators consist of the following layers: C64, C128, C256, C512.

\subsection{Training}
The full loss function combines the reconstruction loss, adversarial loss, domain loss and smoothness loss, denoted respectively as $\Lagr_{\text{cyc}}$, $\Lagr_{\text{adv}}$, $\Lagr_{\text{dom}}$ and $\Lagr_{\text{smooth}}$: 

\begin{equation}
\begin{aligned}
    \Lagr( &G, F, D_X, D_Y, D_H, h ) = \gamma\Lagr_{\text{cyc}}(G, F, h) +
    \epsilon[\Lagr_{\text{adv}}(G, D_Y , X, Y, h ) + 
    \\
    &\Lagr_{\text{adv}}(F, D_X, Y, X, h)] +
    \delta\Lagr_{\text{dom}}(G, F, D_H , X, Y, h ) + \zeta\Lagr_{\text{smooth}}(G, F, h)
\end{aligned}
\end{equation}


We empirically define the values for the weights in the full objective function as: $\gamma = 10$, $\epsilon = 1$, $\delta = 1$, $\zeta=2.5$.
\paragraph{Cycle-consistency loss.}
Rather than enforcing cycle consistency between the input and output images, as in CycleGAN, we enforce it locally along every hop of our multi-hop translation.
That is, $F$ should undo a single hop of $G$ and vice versa.
We enforce this property via a loss proportional to the difference between $F(G_n)$ and $G_{n-1}$ for all hops $n$ (and symmetrically between $G(F_n)$ and $F_{n-1}$:
\begin{equation}
\begin{aligned}
    \Lagr_{\text{cyc}}(&G, F, h) =
    \mathbb{E}_{x \sim p_{\text{data}}(X)} \left[ \sum_{n=1}^{h} || F(G_{n}(x)) - G_{n-1}(x) ||_1\right] +
    \\
    &\mathbb{E}_{y \sim p_{\text{data}}(Y)} \left[\sum_{n=1}^{h} || G(F_{n}(y)) - F_{n-1}(y) ||_1\right]
\end{aligned}
\end{equation}
\paragraph{Adversarial loss.}
The generator $G$ tries to produce images $G_n(x)$ that look similar to those from domain $Y$, while the discriminator $D_Y$ aims to distinguish between the generated images and real images $y \in Y$.
Note that the ``generated images'' include both the final output images and the in-between images. The discriminators use a least squares formulation~\cite{LSGAN}:
\begin{equation}
\begin{aligned}
    &\Lagr_{\text{adv}}(G, D_Y , X, Y, h) = 
    \\
    \mathbb{E}_{y \sim p_{\text{data}}(Y)} & \left[ (D_Y(y) - 1)^2 \right] +
    \mathbb{E}_{x \sim p_{\text{data}}(X)} \left[ \sum_{n=1}^{h} D_Y(G_n(x))^2\right]
\end{aligned}
\end{equation}
\paragraph{Hybrid loss.}
The \textit{hybrid term} assesses the degree to which an image belongs to one of the two domains.
For instance, if \ganh~is trained with $4$ hops, we desire that the first hop $G_1(x)$ be judged as belonging 25\% to domain $Y$ and 75\% to domain $X$.
Thus, we define the \emph{target hybridness score} of hop $G_n$ to be $n/h$; conversely, it is defined as $(h-n)/h$ for the reverse hops $F_n$.
To encourage each hop to achieve its target hybridness, we penalize the distance between the target hybridness and the output of the hybrid discriminator $D_H$ on that hop.
Since $D_H$ is also trained to output 0 for ground-truth images in $X$ and 1 for ground-truth images in $Y$ (i.e. it is a binary domain classifier), an image $i$ for which $D_H(i)$ produces an output of 0.25 can be interpreted as an image which the classifier is 25\% confident belongs to domain $Y$:

\begin{equation}
\begin{aligned}
    \Lagr_{\text{dom}}(&G, F, D_H , X, Y, h ) =
  \mathbb{E}_{x \sim p_{\text{data}}(X)} \left[ \sum_{n=1}^{h} \left( D_H(G_n(x)) - \frac{n}{h} \right)^2 \right] +
  \\
  &\mathbb{E}_{y \sim p_{\text{data}}(Y)} \left[ \sum_{n=1}^{h} \left( D_H(F_n(y)) - \frac{h - n}{h} \right)^2 \right]
\end{aligned}
\end{equation}

\paragraph{Smoothness loss.}
The smoothness term penalizes the image-space difference between hop $n$ and hop $n-1$.
This term encourages the hops to be individually as small as possible while still leading to a full translation when combined, which has a regularizing effect on the training:
\begin{equation}
\begin{aligned}
  \Lagr_{\text{smooth}}(&G, F, h) = 
    \mathbb{E}_{x \sim p_{\text{data}}(X)} \left[ \sum_{n=1}^{h}||{G_{n}(x) - G_{n-1}(x)})||_1\right] +
    \\
    &\mathbb{E}_{y \sim p_{\text{data}}(Y)} \left[\sum_{n=1}^{h}||{F_{n}(y) - F_{n-1}(y)}||_1\right]
\end{aligned}
\end{equation}
\paragraph{Training procedure.}
Algorithm~\ref{alg:training} shows how we train \ganh, i.e. for each image to be translated, we perform a single hop, update the weights of the generator and discriminator networks, perform the next hop, etc.
Training the network this way, rather than performing all hops and then doing a single weight update, has the advantage of requiring significantly less memory.
\wm{Using one specialized generator for each specific hop would considerably increase memory usage, scaling linearly on the number of hops.} 
The \textbf{generator\_update} and \textbf{discriminator\_update} procedures use a single term of the sums which define the loss $\Lagr$ (i.e. the term for hop $n$) to compute parameter gradients.

\begin{figure}[ht]
  \centering
  \begin{minipage}{.7\linewidth}
    \begin{algorithm}[H]
      \SetAlgoLined
      initialize $G$, $F$, $D_X$, $D_Y$, $D_H$\\
      
      \For{each epoch}{
    $x, y \leftarrow$ random\_batch()\;
    $x_{\text{real}}, y_{\text{real}} \leftarrow x, y$ \;
    
    \For{$n = 1$ \normalfont{to} $h$}
   {
     $x, y \leftarrow G(x), F(y)$ \;
     
     \textbf{generator\_update}($G$, $x$, $n$)\;
     
     \textbf{generator\_update}($F$, $y$, $n$)\;
     
     \textbf{discriminator\_update}($D_X$, $y$, $x_{\text{real}}$, $n$)\;
     
     \textbf{discriminator\_update}($D_Y$, $x$, $y_{\text{real}}$, $n$)\;
     
     \textbf{classifier\_update}($D_H$, $x_{\text{real}}$, $y_{\text{real}}$)\;
   }
  }
      
      \caption{Training \ganh}
      \label{alg:training}
    \end{algorithm}
  \end{minipage}
\end{figure}
\section{Results and Evaluation}
\label{sec:results}

Our network takes 128$\times$128 images as input and outputs images of the same resolution.
Experiments were performed on a NVIDIA GTX 1080 Ti (using batch size 6) and a NVIDIA Titan X (batch size 24).
We trained \ganh~using Adam with a learning rate of $0.0002$.
With the exception of the cat/human faces experiment, we trained all experiments for 100 epochs (cat/human mode collapsed after 25 epochs, so we report the results from epoch 22).

In our experiments, we used combinations of seven datasets, translating between pairs. Some translation pairs demand both geometric and texture changes:

\begin{itemize}
    \setlength\itemsep{0pt}
    \item 8,223 dog faces from the Columbia dataset~\cite{columbia_dataset}
    \item 47,906 cat faces from Flickr100m~\cite{DBLP:journals/corr/NiPBEBCW15}
    \item 202,599 human faces from aligned CelebA~\cite{DBLP:journals/corr/LiuLWT14}
    \item The zebra, horse, summer, and winter datasets used in CycleGAN~\cite{zhu2017CycleGAN}
\end{itemize}

We compare \ganh~with four prior approaches: CycleGAN~\cite{zhu2017CycleGAN}, DiscoGAN~\cite{DiscoGAN}, GANimorph~\cite{Gokaslan2018} and UNIT~\cite{UNIT}.
All four are ``unsupervised direct image-to-image translation'' methods, in that they transform the input image from one domain into the output image from another domain without any prior pairing of samples between the two domains.
We trained these baselines on the aforementioned datasets with their public implementation and default settings.


\begin{figure}[!t]
\centering
	\includegraphics[width=0.9\linewidth]{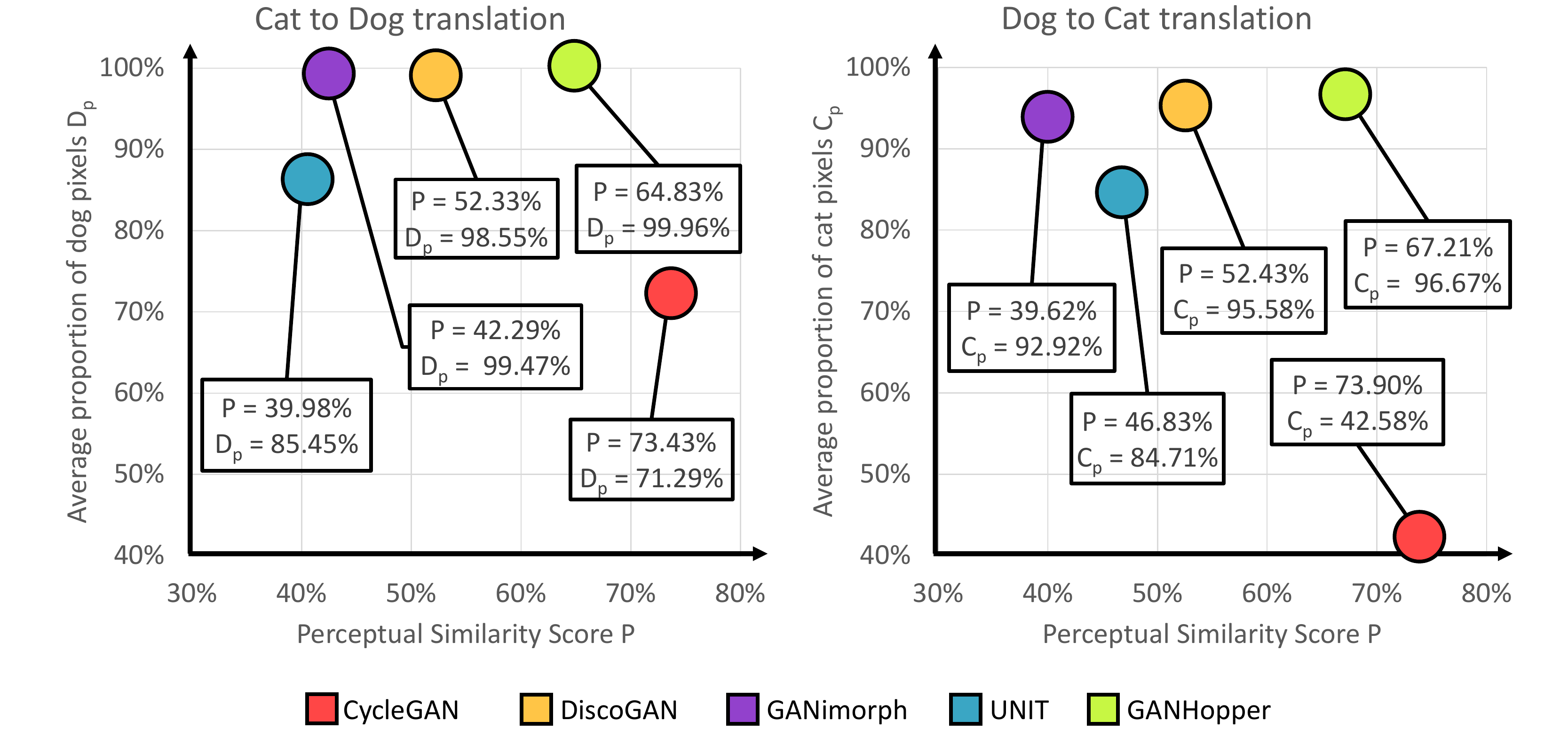}
	\caption{
	Quantitative analysis of dog/cat translation. \ganh~was trained using $4$ hops. The horizontal axis presents the average perceptual similarity~\cite{PerceptualSimilarity} between all inputs and the respective outputs. The vertical axis presents the percentage of output pixels correctly labeled as the output class (e.g. dog or cat) by DeepLabV3~\cite{DeepLabV3} trained on pascal PASCAL VOC 2012. Higher and to the right is better.
}
	\label{fig:quantitative_analysis}
\end{figure}

\begin{figure}[!t]
\centering
	\includegraphics[width=0.85\linewidth]{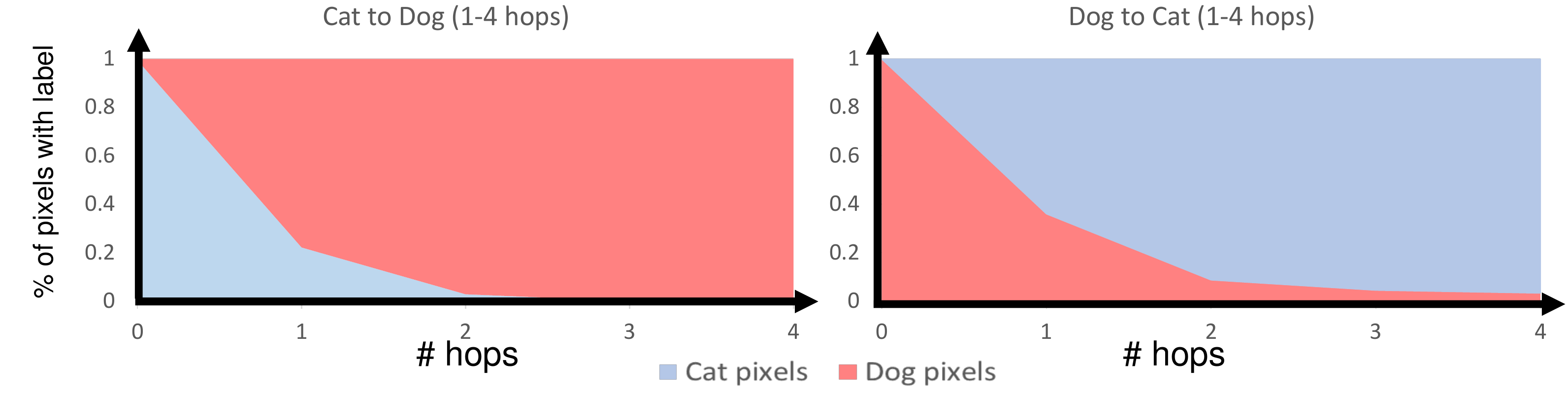}
	\caption{The average percentage of pixels classified as cat or dog (vertical axis) on each hop (horizontal axis).
	\ganh~was trained to translate cats to dogs (and vice versa) using $4$ hops. Pixels classified with any label other than cat or dog are omitted. The $0^{th}$ hop corresponds to the raw inputs. The classification was performed using DeepLabV3~\cite{DeepLabV3} trained on the PASCAL VOC 2012 dataset.}
	\label{fig:graph-hops}
\end{figure}

\wm{
While FID~\cite{FID} is a popular GAN evaluation metric, it is {\em not\/} a specific measure for the {\em translation\/} task.
This is the reason why seminal papers that tackled this task~\cite{zhu2017CycleGAN,DualGAN,Gokaslan2018} did not report FID or other similar metrics.
For instance, CycleGAN performed their quantitative evaluation by quantifying the proportion of pixels in Cityscape's label-to-photo translation correctly classified by a pretrained FCN~\cite{FCN} (Fully Connected Networks) using the input label image as the ground truth for the metric.
Similar to \ganh, GANimorph proposed to evaluate image-to-image translations using two complimentary measures: DeepLabV3's pixel-wise labeling~\cite{DeepLabV3} and a perceptual similarity metric~\cite{PerceptualSimilarity}. 
Therefore, we first compute the percentage of output pixels that are classified as belonging to the target domain by a pre-trained semantic segmentation network (DeepLabV3~\cite{DeepLabV3}, trained on PASCAL VOC 2012).
Secondly, we measure how well the output preserves salient features from the input using a perceptual similarity metric~\cite{PerceptualSimilarity}.
We argue that the combination of these two metrics provide a better evaluation for image-to-image translation networks.

The results of this quantitative analysis can be seen in Figure~\ref{fig:quantitative_analysis}.
CycleGAN produces outputs that best resemble the input but fails to perform domain translation.
Our approach outperforms UNIT, GANimorph and DiscoGAN on both metrics.
This result indicates that one need not necessarily sacrifice domain translation ability to preserve salient features of the input.
}
Figure~\ref{fig:graph-hops} shows how the percentage of pixels translated varies as a function of the number of hops performed.
While not strictly linearly translating the pixels, it is still a smooth monotonic function, suggesting that hops successfully encourage in-between images that can be interpreted as domain hybrids. 

\begin{table}
\begin{center}
\caption{Qualitative analysis of cat-to-dog and dog-to-cat translations performed using Amazon Mechanical Turk to evaluate how real the generated images are perceived to be and the extent to which \ganh~and the image-to-image translation baselines succeed at the translation task}
\label{tab:perceptual_study}
\begin{tabular}{r|cc|cc|}
\cline{2-5}
\multicolumn{1}{l|}{}                      & \multicolumn{2}{c|}{Cat-to-dog}     & \multicolumn{2}{c|}{Dog-to-cat}                                  \\ \hline
\multicolumn{1}{|c|}{Approach}             & Real/Fake        & Translation      & \multicolumn{1}{l}{Real/Fake} & \multicolumn{1}{l|}{Translation} \\ \hline
\multicolumn{1}{|r|}{\ganh} & 26.98\%          & \textbf{38.11\%} & \textbf{36.94\%}              & \textbf{63.76\%}                 \\
\multicolumn{1}{|r|}{UNIT}                 & 20.08\%          & 35.19\%          & 31.47\%                       & 16.28\%                          \\
\multicolumn{1}{|r|}{DiscoGAN}             & 24.73\%          & 20.61\%          & 29.55\%                       & 12.76\%                          \\
\multicolumn{1}{|r|}{GANimorph}           & \textbf{36.24\%} & 6.07\%           & 29.61\%                       & 7.2\%                            \\ \hline
\end{tabular}
\end{center}
\end{table}

\wm{Perceptual studies are the gold
standard for assessing graphical realism~\cite{zhu2017CycleGAN}. Therefore, we performed a perceptual evaluation to measure (a) the extent to which dog and cat faces generated by our method, DiscoGAN, GANimorph, and UNIT} are perceived as real; and (b) how much the generated samples from these four approaches resemble the input image.
\wm{
For evaluation (a), 36 participants were exposed to 128 paired random samples of cat-to-dog translations while 36 other participants were exposed to the same amount of random dog-to-cat translations.
}
Both images of each pair were shown for one second at the same time. Also, for each given pair, one of the images was always a real sample while the other was a fake generated by one of the image translation networks. After an image pair was shown, the participant was asked which one of the two images shown looked real. A score of 50\% for a given method would indicate that participants were unable to discriminate between real data and generated data on a given domain.
For evaluation (b), we displayed 128 random images from our input test data to the participants, and the outputs of the translation using \wm{UNIT, DiscoGAN, GANimorph and \ganh}. 
Afterwards, the participant was asked which image better translates the input to the target domain. For (b) there was no time constraint and, as in (a), 18 participants were evaluated for each class. Therefore, the perceptual study with humans had 108 participants in total.
\wm{
As shown in Table~1, the perceptual experiment results indicate that our method was outperformed exclusively by GANimorph in how real the generated dogs are perceived to be. However, \ganh~outperforms GANimorph if one also takes into account the translation task, measure in which the later network had the worst performance compared to the other approaches. Furthermore, \ganh~significantly outperforms all other three methods in how real generated cats are perceived to be by humans and also on the translation task for both dog-to-cat and cat-to-dog translations.}

\begin{figure*}[!t]
\centering
	\includegraphics[width=0.99\textwidth]{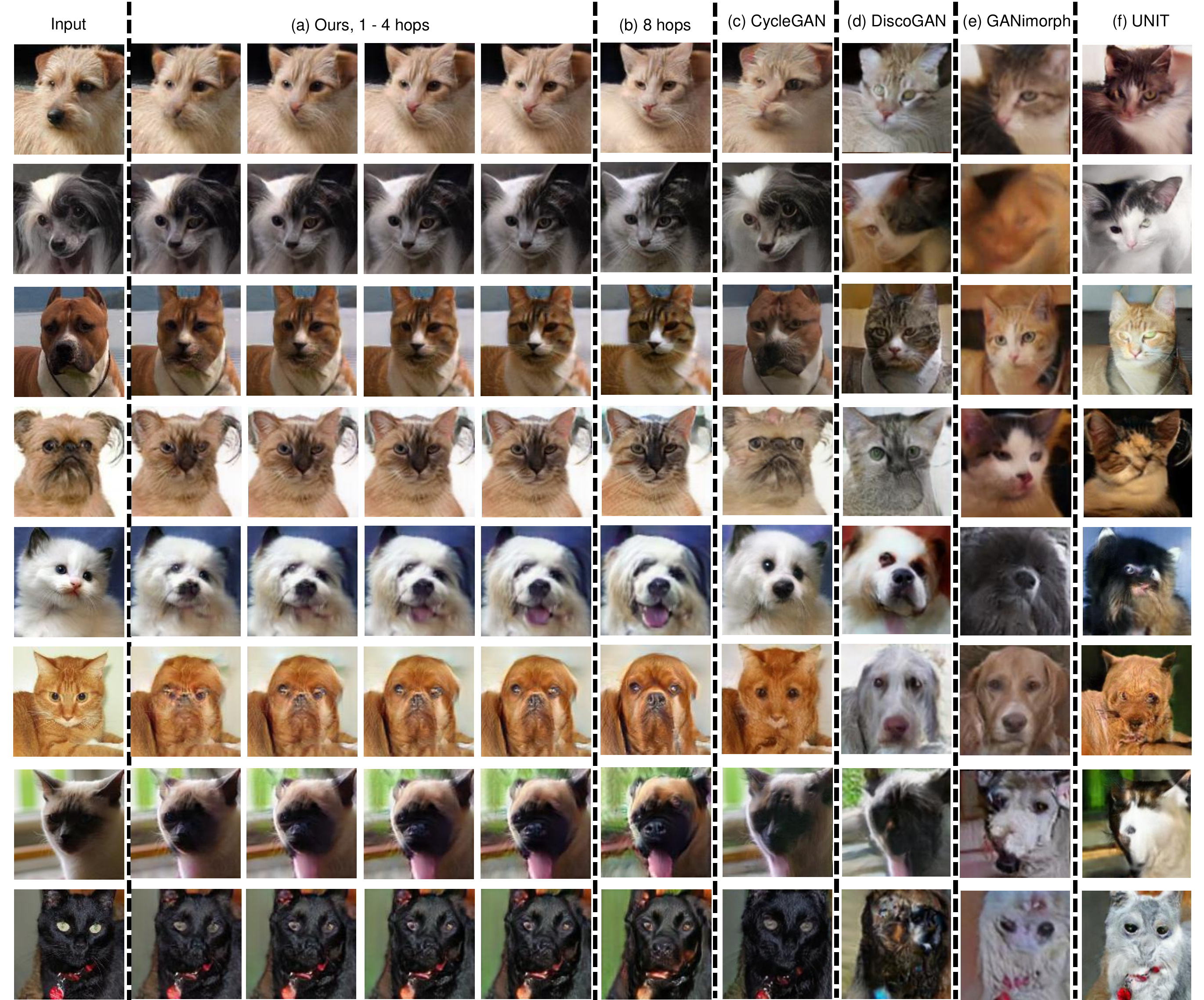}
	\caption{Comparing different translation methods on the challenging dog/cat faces dataset. We trained \ganh~with four hops; (a) shows the result of hopping 1 to 4 times from the input and (b) shows the result of 8 hops from the input. We compare our results to (c) CycleGAN, (d) DiscoGAN, (e) GANimorph, and (f) UNIT.
	}
	\label{fig:3-hops_dog_cat_2.5}
\end{figure*}

Figure~\ref{fig:3-hops_dog_cat_2.5} compares our method to the baselines on cat to dog and dog to cat translation.
Our multi-hop procedure translates the input via a sequence of hybrid images (Figure~\ref{fig:3-hops_dog_cat_2.5}(a)), allowing it to preserve key visual characteristics of the input if changing them is not necessary to achieve domain translation.
For instance, fur colors and background textures are preserved in most cases (e.g. white cats map to white dogs) as is head orientation, while domain-specific features such as eyes, noses, and ears are appropriately deformed.
The multi-hop procedure also allows control over how much translation to perform.
The user can control the degree of ``dogness'' or ``catness'' introduced by the translation, including performing more hops than the network was trained on in order to exaggerate the characteristics of the target domain.
Figure~\ref{fig:3-hops_dog_cat_2.5}(b) shows the result of performing 8 hops using a network trained to perform a complete translation using 4 hops.
Note that, in the fifth row of Figure~\ref{fig:3-hops_dog_cat_2.5}, the additional hops help to clarify the shape of the output dog's tongue.

By contrast, the baselines produce less desirable results.
CycleGAN preserves the input features too much, leading to incomplete translations (Figure~\ref{fig:3-hops_dog_cat_2.5}(c)).
Note that CycleGAN's outputs often look similar to the first hop of our network; this makes sense, since each hop uses a CycleGAN-like generator network.
Our network uses multiple hops of that same architecture to overcome CycleGAN's original limitations.
DiscoGAN (Figure~\ref{fig:3-hops_dog_cat_2.5}(d)) can properly translate high-level properties such as head pose and eye placement but fails to preserve lower-level appearance details such as fur patterns and color.
Its results are also often geometrically malformed (lines 2, 4, 5, 7, and 8 on Figure~\ref{fig:3-hops_dog_cat_2.5}).
GANimorph (Figure~\ref{fig:3-hops_dog_cat_2.5}(e)) produces images that are convincingly part of the target domain but preserve little of the input image's features (typically only head pose).
\wm{
Finally, while UNIT (Figure~\ref{fig:3-hops_dog_cat_2.5}(f)) produces images that normally preserve head pose, features like fur patterns are needlessly changed in the translation process. For instance, a white cat should be translated to a white dog instead of a black dog, as shown on line 5.
Note that all networks besides \ganh~and CycleGAN tend to produce outputs with noticeably decreased saturation and contrast.
}

\begin{figure*}[!t]
\centering
	\includegraphics[width=0.99\textwidth]{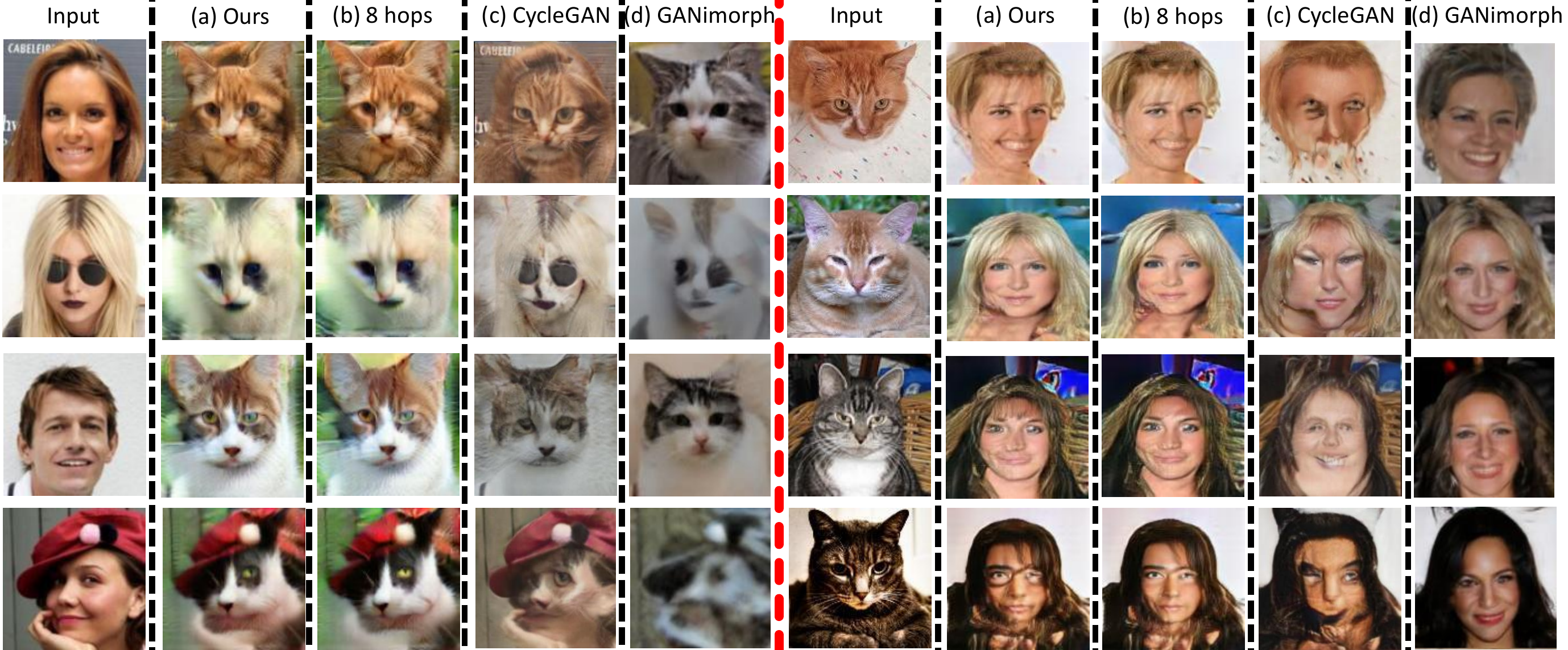}
	\caption{Examples of human-to-cat faces translation. The approaches compared are (a) \ganh, (b) $4$ extra hops after the full translation, (c) CycleGAN and (d) GANimorph.
	\ganh~was trained with default hyperparameter values.
	}
	\label{fig:3-hops_human_cat_2.5}
\end{figure*}


Figure~\ref{fig:3-hops_human_cat_2.5} shows a similar comparison on human to cat translation.
Again, our method preserves input features well: facial structures stay roughly the same, and cats with light fur tend to generate blonde-haired people.
Our method also preserves background details better than the baselines.


\begin{figure}[!t]
    \centering
	\includegraphics[width=0.99\linewidth]{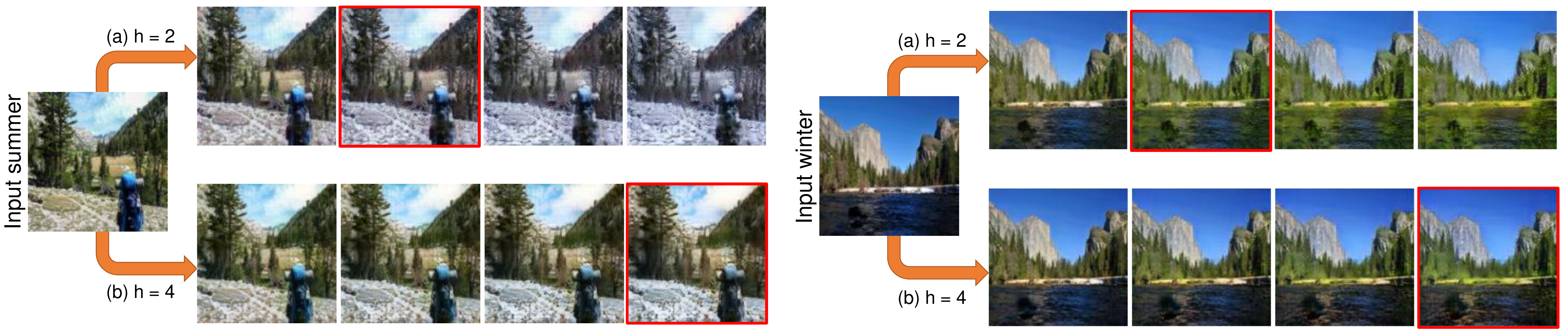}
	\caption{Impact of training hop count.
	Using $h = 4$ hops (b) better preserves input features, but using $h = 2$ hops (a) allows more drastic changes. Red squares denote the hops that correspond to a full translation in each setting; images further to the right are extrapolations obtained by applying additional hops.
	}
	\label{fig:summer2winter-hop-comparison}
\end{figure}

\begin{figure}[!t]
	\includegraphics[width=0.69\linewidth]{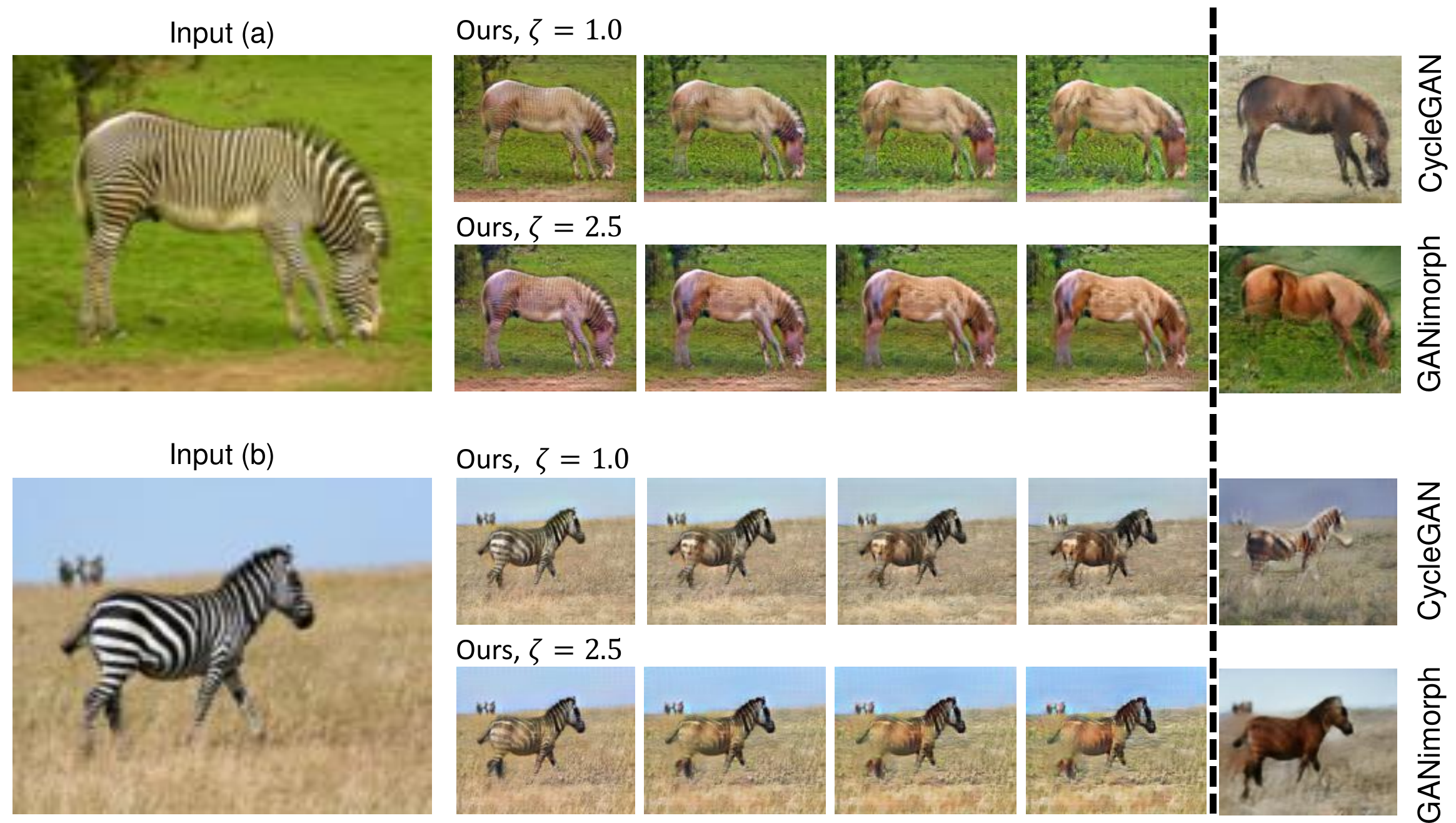}
	\centering
	\caption{
	\wm{As with CycleGAN and GANimorph, our method occasionally ``erases'' part of an object and replaces it with background, rather than correctly translating it (e.g. the zebra legs disappear).
	This can be mitigated by increasing the smoothness loss weight $\zeta$.
	All other hyperparameters are set to default values.
	}
	}
	\label{fig:horse2zebra}
\end{figure}

We also examine the impact of the number of hops used during training.
A network using too few hops must more quickly change the domain of the image; this causes the generator to ``force'' the translation and produce undesirable outputs.
In the summer to winter translation of Figure~\ref{fig:summer2winter-hop-comparison} (Top), the hiker's jacket quickly loses its blue color in the first row ($h = 2$) compared with the second row ($h = 4$).
In the winter to summer translation of Figure~\ref{fig:summer2winter-hop-comparison} (Bottom), the lake incorrectly becomes green when using a two-hop network but is preserved with four hops (while vegetation is still converted to green).
The results suggest that increasing the number of hops has the added benefit of increasing image diversity and also allowing for smoother transition from one domain to another.
\wm{
Figure~\ref{fig:summer2winter-hop-comparison} and Figure~\ref{fig:horse2zebra} also show how \ganh~addresses datasets with varying dominant color schemes on each domain: colors are smoothly interpolated from the input to the output on each hop until the translation process terminates.
}

\begin{figure}[!t]
    \centering
	\includegraphics[width=\linewidth]{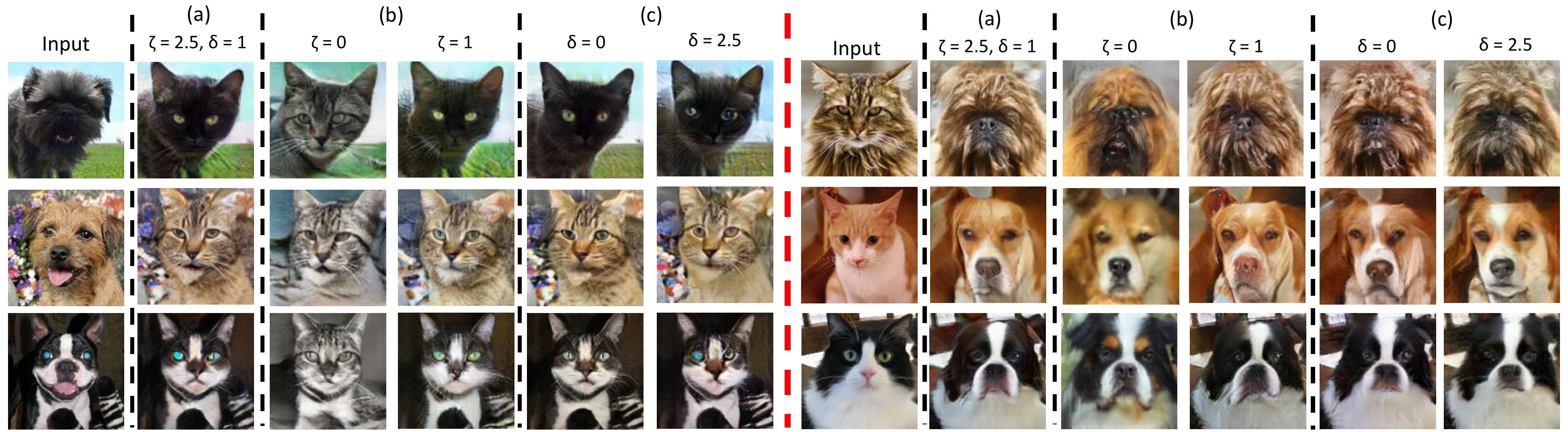}
	\caption{
	\wm{Evaluation of the impact of the smoothness term weight $\zeta$ and hybrid term $\delta$ on the dog to cat dataset trained with $4$ hops. The results using default hyperparameters are shown in (a). The value of $\delta$ at (b) is $1$ and the value of $\zeta$ at (c) is $2.5$.}
	}
	\label{fig:impact_of_smoothness}
\end{figure}

\wm{
Figure~\ref{fig:impact_of_smoothness} demonstrates the impact of the smoothness weight $\zeta$ on training dog-to-cat translations with 4 hops. Default hyperparameters help \ganh~to preserve the original fur patterns while still preserving sharp local features in both translation directions, as shown in Figure~\ref{fig:impact_of_smoothness}(a). With $\zeta = 0$ , as shown in Figure~\ref{fig:impact_of_smoothness}(b)(Left), the network collapses to producing mostly cats with gray and white fur, while generating noticeably blurrier dogs. Figure~\ref{fig:impact_of_smoothness}(b)(Right) shows that, as $\zeta$ increases, both issues are progressively mitigated.
Further, we can observe in Figure~\ref{fig:impact_of_smoothness}(c)(Left) that $\delta = 0$ tends to produce slightly less sharp features. This effect is more pronounced in the cat-to-dog translation than in the reverse direction. Figure~\ref{fig:impact_of_smoothness}(c)(Right) shows that increasing $\delta$ to $2.5$ leads to more artifacts in the translation process. For instance, note the asymmetry in the left and right eyes of the dog-to-cat translations. }



As our method uses CycleGAN as a sub-component, it inherits some of the problems faced by that method, as well as other direct unsupervised image translators.
Figure~\ref{fig:horse2zebra} shows one prominent failure mode, in which the network ``cheats'' by erasing part of the object to be translated and replacing it with background (e.g. zebra legs).
The smoothness term in our loss function penalizes differences between hops, so increasing its weight mitigates this issue.





\section{Conclusion and Future Work}
\label{sec:future}

Unsupervised image-to-image translation is an ill-posed problem. Different methods have chosen different regularizing assumptions to drive their solutions~\cite{TransGaGa,UNIT,MUNIT}.
In this paper, we follow the cycle-consistency assumption of CycleGAN~\cite{zhu2017CycleGAN} and DualGAN~\cite{DualGAN}, while introducing the multi-hop paradigm to exert fine-grained control over the translation using a new hybrid discriminator.
\rz{As shown by both the quantitative analysis and human evaluation experiment presented in Section~\ref{sec:results}, \ganh~outperforms other baseline approaches by better preserving features of the input images while still applying the necessary transformations to create outputs which clearly belong to the target domain.}



The meta idea of ``transforming images in small steps'' raises new and interesting questions worth exploring.
For example, how many steps are ideal?
The results in this paper used 2-4 hops, as more hops did not noticeably improve performance but did increase training time.
However, some images in a domain $X$ are clearly harder than others to translate into a different domain $Y$ (e.g. translating dogs with long vs. short snouts into cats).
Can we automatically learn the ideal number of hops for each input image?
Taken to an extreme, can we use a very large number of tiny hops to produce a smooth interpolation sequence from source to target domain?
We also want to identify domains where \ganh~systematically fails and explore the design space of multi-hop translation architectures in response.
For instance, while \ganh~uses the same network for all hops, it may be better to use different networks per hop (i.e. the optimal function for translating a 25\% dog to a 50\% dog may not be the same as the function for translating a 75\% dog to a real dog).

It would be interesting to combine \ganh~with ideas from MUNIT~\cite{MUNIT} or BiCycleGAN~\cite{BicycleGAN}, so that the user can control the output of the translation via a ``style'' code while still preserving important input features (e.g. translating a white cat into different white-furred dog breeds).
\rz{Yet another potential future work is to include the current hop information as part of the generator input (e.g.~one-hot vector) to avoid the reliance on the generator to infer which hop operation should be performed based only on the input image.}
Finally, we would like to investigate the idea that initially spurred the development of \ganh: generating meaningful \emph{extrapolation} sequences \emph{beyond} the boundaries of a given image domain, to produce creative and novel outputs.


%
%
%
\bibliographystyle{splncs04}
\bibliography{hopgan}

\begin{thebibliography}{10}
\providecommand{\url}[1]{\texttt{#1}}
\providecommand{\urlprefix}{URL }
\providecommand{\doi}[1]{https://doi.org/#1}

\bibitem{NeuralBestBuddies}
Aberman, K., Liao, J., Shi, M., Lischinski, D., Chen, B., Cohen-Or, D.: Neural
  best-buddies: Sparse cross-domain correspondence. ACM Trans. Graph.
  \textbf{37}(4) (Jul 2018)

\bibitem{CariGAN}
Cao, K., Liao, J., Yuan, L.: Carigans: Unpaired photo-to-caricature translation
  (2018)

\bibitem{DeepLabV3}
Chen, L., Papandreou, G., Schroff, F., Adam, H.: Rethinking atrous convolution
  for semantic image segmentation. CoRR  \textbf{abs/1706.05587} (2017)

\bibitem{InfoGAN}
Chen, X., Duan, Y., Houthooft, R., Schulman, J., Sutskever, I., Abbeel, P.:
  Infogan: Interpretable representation learning by information maximizing
  generative adversarial nets. In: Proceedings of the 30th International
  Conference on Neural Information Processing Systems (2016)

\bibitem{Gokaslan2018}
Gokaslan, A., Ramanujan, V., Ritchie, D., Kim, K.I., Tompkin, J.: Improving
  shape deformation in unsupervised image-to-image translation. CoRR
  \textbf{abs/1808.04325} (2018), \url{http://arxiv.org/abs/1808.04325}

\bibitem{FID}
Heusel, M., Ramsauer, H., Unterthiner, T., Nessler, B., Hochreiter, S.: Gans
  trained by a two time-scale update rule converge to a local nash equilibrium.
  In: Guyon, I., Luxburg, U.V., Bengio, S., Wallach, H., Fergus, R.,
  Vishwanathan, S., Garnett, R. (eds.) Advances in Neural Information
  Processing Systems 30, pp. 6626--6637. Curran Associates, Inc. (2017),
  \url{http://papers.nips.cc/paper/7240-gans-trained-by-a-two-time-scale-update-rule-converge-to-a-local-nash-equilibrium.pdf}

\bibitem{MUNIT}
Huang, X., Liu, M.Y., Belongie, S., Kautz, J.: Multimodal unsupervised
  image-to-image translation. In: ECCV (2018)

\bibitem{IsolaZZE16}
Isola, P., Zhu, J., Zhou, T., Efros, A.A.: Image-to-image translation with
  conditional adversarial networks. CoRR  \textbf{abs/1611.07004} (2016),
  \url{http://arxiv.org/abs/1611.07004}

\bibitem{JohnsonAL16}
Johnson, J., Alahi, A., Li, F.: Perceptual losses for real-time style transfer
  and super-resolution. CoRR  \textbf{abs/1603.08155} (2016),
  \url{http://arxiv.org/abs/1603.08155}

\bibitem{CascadedDeepFeatureTranslation}
Katzir, O., Lischinski, D., Cohen{-}Or, D.: Cross-domain cascaded deep feature
  translation. In: European Conference on Computer Vision (ECCV). Springer
  (2020)

\bibitem{DiscoGAN}
Kim, T., Cha, M., Kim, H., Lee, J.K., Kim, J.: Learning to discover
  cross-domain relations with generative adversarial networks. In: ICML (2017)

\bibitem{FaderNetworks}
Lample, G., Zeghidour, N., Usunier, N., Bordes, A., DENOYER, L., et~al.: Fader
  networks: Manipulating images by sliding attributes. In: Advances in Neural
  Information Processing Systems (2017)

\bibitem{AutomatingImageMorphing}
Liao, J., Lima, R.S., Nehab, D., Hoppe, H., Sander, P.V., Yu, J.: Automating
  image morphing using structural similarity on a halfway domain. ACM Trans.
  Graph.  \textbf{33}(5) (Sep 2014)

\bibitem{columbia_dataset}
Liu, J., Kanazawa, A., Jacobs, D., Belhumeur, P.: Dog breed classification
  using part localization. In: Fitzgibbon, A., Lazebnik, S., Perona, P., Sato,
  Y., Schmid, C. (eds.) Computer Vision -- ECCV 2012. pp. 172--185. Springer
  Berlin Heidelberg, Berlin, Heidelberg (2012)

\bibitem{UNIT}
Liu, M., Breuel, T., Kautz, J.: Unsupervised image-to-image translation
  networks. CoRR  \textbf{abs/1703.00848} (2017),
  \url{http://arxiv.org/abs/1703.00848}

\bibitem{DBLP:journals/corr/LiuLWT14}
Liu, Z., Luo, P., Wang, X., Tang, X.: Deep learning face attributes in the
  wild. CoRR  \textbf{abs/1411.7766} (2014),
  \url{http://arxiv.org/abs/1411.7766}

\bibitem{LSGAN}
Mao, X., Li, Q., Xie, H., Lau, R.Y.K., Wang, Z.: Least squares generative
  adversarial networks. In: ICCV (2017)

\bibitem{instagan}
Mo, S., Cho, M., Shin, J.: Instagan: Instance-aware image-to-image translation.
  In: International Conference on Learning Representations (2019),
  \url{https://openreview.net/forum?id=ryxwJhC9YX}

\bibitem{DBLP:journals/corr/NiPBEBCW15}
Ni, K., Pearce, R.A., Boakye, K., Essen, B.V., Borth, D., Chen, B., Wang, E.X.:
  Large-scale deep learning on the {YFCC100M} dataset. CoRR
  \textbf{abs/1502.03409} (2015), \url{http://arxiv.org/abs/1502.03409}

\bibitem{UNet}
Ronneberger, O., Fischer, P., Brox, T.: U-net: Convolutional networks for
  biomedical image segmentation. In: MICCAI (2015)

\bibitem{FCN}
Shelhamer, E., Long, J., Darrell, T.: Fully convolutional networks for semantic
  segmentation. IEEE Transactions on Pattern Analysis and Machine Intelligence
  \textbf{39}(4),  640--651 (April 2017). \doi{10.1109/TPAMI.2016.2572683},
  \url{http://ieeexplore.ieee.org/document/7478072/}

\bibitem{DTN}
Taigman, Y., Polyak, A., Wolf, L.: Unsupervised cross-domain image generation.
  In: Proc. of ICLR (2017)

\bibitem{Pix2PixHD}
Wang, T.C., Liu, M.Y., Zhu, J.Y., Tao, A., Kautz, J., Catanzaro, B.:
  High-resolution image synthesis and semantic manipulation with conditional
  gans. In: Proceedings of the IEEE Conference on Computer Vision and Pattern
  Recognition (2018)

\bibitem{TransGaGa}
Wu, W., Cao, K., Li, C., Qian, C., Loy, C.C.: Transgaga: Geometry-aware
  unsupervised image-to-image translation. In: Proc. of CVPR (2019)

\bibitem{DualGAN}
Yi, Z., Zhang, H., Tan, P., Gong, M.: {DualGAN}: Unsupervised dual learning for
  image-to-image translation. In: Proc. of ICCV (2017)

\bibitem{LOGAN}
Yin, K., Chen, Z., Huang, H., Cohen-Or, D., Zhang, H.: {LOGAN}: Unpaired shape
  transform in latent overcomplete space. ACM Trans. on Graphics
  \textbf{38}(6) (2019)

\bibitem{PerceptualSimilarity}
{Zhang}, R., {Isola}, P., {Efros}, A.A., {Shechtman}, E., {Wang}, O.: The
  unreasonable effectiveness of deep features as a perceptual metric. In: 2018
  IEEE/CVF Conference on Computer Vision and Pattern Recognition. pp. 586--595
  (June 2018). \doi{10.1109/CVPR.2018.00068}

\bibitem{zhu2017CycleGAN}
Zhu, J., Park, T., Isola, P., Efros, A.A.: Unpaired image-to-image translation
  using cycle-consistent adversarial networks. In: International Conference on
  Computer Vision (ICCV), to appear (2017)

\bibitem{BicycleGAN}
Zhu, J.Y., Zhang, R., Pathak, D., Darrell, T., Efros, A.A., Wang, O.,
  Shechtman, E.: Toward multimodal image-to-image translation. In: Advances in
  Neural Information Processing Systems (2017)

\end{thebibliography}
\end{document}